\documentclass{article}

\PassOptionsToPackage{numbers, compress}{natbib}
\usepackage[preprint]{neurips_2020}

\usepackage[utf8]{inputenc} % allow utf-8 input
\usepackage[T1]{fontenc}    % use 8-bit T1 fonts
\usepackage[colorlinks=True]{hyperref}  % hyperlinks
\usepackage{url}            % simple URL typesetting
\usepackage{booktabs}       % professional-quality tables
\usepackage{amsfonts}       % blackboard math symbols
\usepackage{nicefrac}       % compact symbols for 1/2, etc.
\usepackage{microtype}      % microtypography

\usepackage{bold-extra}
\usepackage[dvipsnames]{xcolor}
\usepackage{multirow}
\usepackage{amssymb}
\usepackage{makecell}

\usepackage{epsfig}
\usepackage{graphicx}
\usepackage{amsmath}
\usepackage{multirow}

\usepackage{caption}
\usepackage{subcaption}

\usepackage[ruled]{algorithm2e}
\usepackage[dvipsnames]{xcolor}
\usepackage{listings}

\usepackage{perpage}
\MakePerPage{footnote}

\title{Self-Supervised Learning with Swin Transformers}

\author{
   Zhenda Xie\thanks{Equal contribution. $^\dag$Interns at MSRA.}~~$^{\dag13}$ Yutong Lin$^{*\dag23}$ Zhuliang Yao$^{\dag13}$ ~Zheng Zhang$^3$ ~Qi Dai$^3$ ~Yue Cao$^3$ ~Han Hu$^3$\\
   $^1$Tsinghua University \quad $^2$Xi'an Jiaotong University \\
   $^3$Microsoft Research Asia \\
   \texttt{\{xzd18,yzl17\}@mails.tsinghua.edu.cn} \quad  \texttt{yutonglin@stu.xjtu.edu.cn} \\
   \texttt{\{zhez,qid,yuecao,hanhu\}@microsoft.com}
}

\begin{document}

\maketitle

\begin{abstract}

We are witnessing a modeling shift from CNN to Transformers in computer vision. In this work, we present a self-supervised learning approach called \textcolor{Maroon}{\texttt{MoBY}}, with Vision Transformers as its backbone architecture. The approach basically has no new inventions, which is combined from \textcolor{Maroon}{\texttt{Mo}}\texttt{Co} \texttt{v2} and \textcolor{Maroon}{\texttt{BY}}\texttt{OL} and tuned to achieve reasonably high accuracy on ImageNet-1K linear evaluation: 72.8\% and 75.0\% top-1 accuracy using DeiT-S and Swin-T, respectively, by 300-epoch training. The performance is slightly better than recent works of MoCo v3 and DINO which adopt DeiT as the backbone, but with much lighter tricks. 

More importantly, the general-purpose Swin Transformer backbone enables us to also evaluate the learnt representations on downstream tasks such as object detection and semantic segmentation, in contrast to a few recent approaches built on ViT/DeiT which only report linear evaluation results on ImageNet-1K due to ViT/DeiT not tamed for these dense prediction tasks. We hope our results can facilitate more comprehensive evaluation of self-supervised learning methods designed for Transformer architectures. Our code and models are available at \url{https://github.com/SwinTransformer/Transformer-SSL}, which will be continually enriched. 

\end{abstract}

\section{Introduction}

The vision field is undergoing two revolutionary trends since about two years ago. The first trend is self-supervised visual representation learning pioneered by MoCo~\citep{moco}, which for the first time demonstrated superior transferring performance on seven downstream tasks over the previous standard supervised methods by ImageNet-1K classification. The second is the Transformer-based backbone architecture~\citep{vit,deit,swin}, which has strong potential to replace the previous standard convolutional neural networks such as ResNet~\citep{resnet}. The pioneer work is ViT~\citep{vit}, which demonstrated strong performance on image classification by directly applying the standard Transformer encoder~\citep{attention} in NLP on non-overlapping image patches. The follow-up work, DeiT~\citep{deit}, tuned several training strategies to make ViT work well on ImageNet-1K image classification. While ViT/DeiT are designed for the image classification task and has not been well tamed for downstream tasks requiring dense prediction, Swin Transformer~\citep{swin} is proposed to serve as a general-purpose vision backbone by introducing useful inductive biases of locality, hierarchy and translation invariance.

While the two revolutionary waves appeared independently, the community is curious about what kind of adaptation is needed and what it will behave when they meet each other. Nevertheless, until very recently, a few works started to explore this space: MoCo v3~\citep{mocov3} presents a training recipe to let ViT perform reasonably well on ImageNet-1K linear evaluation; DINO~\citep{dino} presents a new self-supervised learning method which shows good synergy with the Transformer architecture.

Although these works produce encouraging results on ImageNet-1K linear evaluation, there are no assessment of the transferring performance on downstream tasks such as object detection and semantic segmentation, probably due to that ViT/DeiT are not well tamed for these downstream tasks. To enable more comprehensive evaluations of the self-supervised learnt representations on also these downstream tasks, we propose to adopt Swin Transformer as the backbone architecture instead of the previous used ViT architecture, thanks to that Swin Transformer is designed as general-purpose and performs strong on downstream tasks.

In addition to this backbone architecture change, we also present a self-supervised learning approach by combining MoCo v2~\citep{mocov2} and BYOL~\citep{byol}, named \textcolor{Maroon}{\texttt{MoBY}} (by picking the first two letters of each). We tune a training recipe to make the approach performing reasonably high on ImageNet-1K linear evaluation: 72.8\% top-1 accuracy using DeiT-S with 300-epoch training which is slightly better than that in MoCo v3 and DINO but with lighter tricks. Using Swin-T architecture instead of DeiT-S, it achieves 75.0\% top-1 accuracy with 300-epoch training, which is 2.2\% higher than that using DeiT-S. 
Initial study shows that some tricks in MoCo v3 and DINO are also useful for MoBY, e.g. replacing the LayerNorm layers before the MLP blocks by BatchNorm like that in MoCo v3 bring additional +1.1\% gains using 100 epoch training, indicating the strong potential of MoBY.

When transferred to downstream tasks of COCO object detection and ADE20K semantic segmentation, the representations learnt by this self-supervised learning approach achieves on par performance compared to the supervised method. Noting self-supervised learning with ResNet architectures has shown significantly stronger transferring performance on downstream tasks than supervised methods~\citep{moco,pixpro,henaff2021efficient}, the results indicate large space to improve for self-supervised learning with Transformers.

The proposed approach basically has no new inventions. What we provide is an approach which combines the previous good practice but with lighter tricks, associated with tuned hyper-parameters to achieve reasonably high accuracy on ImageNet-1K linear evaluation. We also provide baselines to aid the evaluation of transferring performance on downstream tasks for the future study of self-supervised learning on Transformer architectures.

\section{A Baseline SSL Method with Swin Transformers}

\paragraph{MoBY: a self-supervised learning approach} MoBY is a combination of two popular self-supervised learning approaches: MoCo v2~\citep{mocov2} and BYOL~\citep{byol}. It inherits the momentum design, the \emph{key} queue, and the contrastive loss used in MoCo v2, and inherits the asymmetric encoders, asymmetric data augmentations and the momentum scheduler in BYOL. We name it \textcolor{Maroon}{\texttt{MoBY}} by picking the first two letters of each method.

\begin{figure}[h]
\centering
\includegraphics[width=1.0\linewidth]{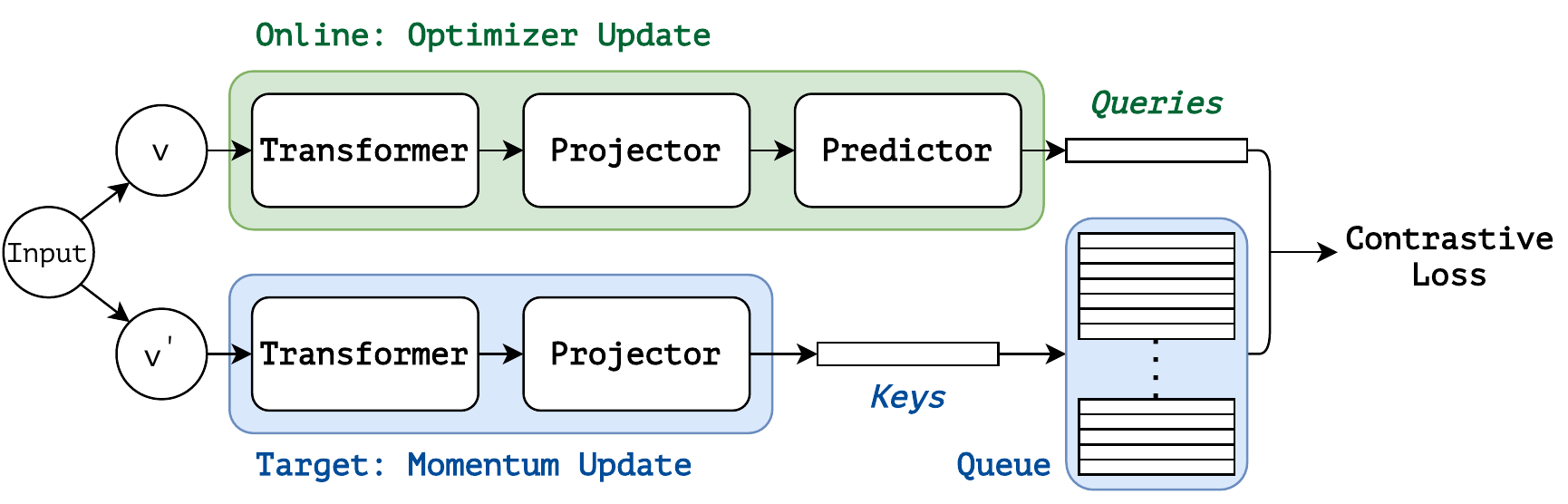}
\caption{The pipeline of MoBY.}
\label{fig-moby}
\end{figure}

The MoBY approach is illustrated in Figure~\ref{fig-moby}. There are two encoders: an \textit{online} encoder and a \textit{target} encoder. Both two encoders consist of a backbone and a projector head (2-layer MLP), and the \textit{online} encoder introduces an additional prediction head (2-layer MLP), which makes the two encoders asymmetric. The \textit{online} encoder is updated by gradients, and the \textit{target} encoder is a moving average of the \textit{online} encoder by momentum updating in each training iteration. A gradually increasing momentum updating strategy is applied for on the \textit{target} encoder: the value of momentum term is gradually increased to 1 during the course of training. The default starting value is 0.99.

A contrastive loss is applied to learn the representations. Specifically, for an \emph{online} view $q$, its contrastive loss is computed as
\begin{equation}
    \mathcal{L}_q = -\text{log} \frac{\text{exp}(q \cdot k_{+} / \tau)}{\sum_{i=0}^{K} \text{exp}(q \cdot k_i / \tau)},
\end{equation}
where $k_{+}$ is the \emph{target} feature for the other view of the same image; $k_i$ is a \emph{target} feature in the \emph{key} queue; $\tau$ is a temperature term; $K$ is the size of the \emph{key} queue (4096 by default).

In training, like most Transformer-based methods, we also adopt the AdamW~\citep{adam,adamw} optimizer, in contrast to previous self-supervised learning approaches built on ResNet backbone where usually SGD~\citep{moco,pic} or LARS~\citep{simclr,byol,pixpro} is used. We also introduce a regularization method of \emph{asymmetric drop path} which proves crucial for the final performance.

In the experiments, we adopt a fixed learning rate of 0.001 and a fixed weight decay of 0.05, which performs stably well. We tune hyper-parameters of the \emph{key} queue size $K$, the starting momentum value of the target branch, the temperature $\tau$, and the drop path rates. 

A pseudo code of MoBY in a PyTorch-like style is shown in Algorithm~\ref{algorithm-moby}.

\definecolor{codeblue}{rgb}{0.25,0.5,0.5}
\definecolor{keyword}{rgb}{0.8, 0.25, 0.5}
\lstset{
  backgroundcolor=\color{white},
  basicstyle=\fontsize{8pt}{8pt}\ttfamily,
  columns=fullflexible,
  breaklines=false,
  commentstyle=\fontsize{8pt}{8pt}\color{codeblue},
  keywordstyle=\fontsize{8pt}{8pt}\color{keyword},
}
\begin{figure}
\centering
\begin{algorithm}[H]
\caption{Pseudo code of MoBY in a PyTorch-like style.}
\label{algorithm-moby}
\begin{lstlisting}[language=python,tabsize=4,showtabs]
# encoder: transformer-based encoder
# proj: projector
# pred: predictor
# odpr: online drop path rate
# tdpr: target drop path rate
# m: momentum coefficient
# t: temperature coefficient
# queue1, queue2: two queues for storing negative samples

f_online = lambda x: pred(proj(encoder(x, drop_path_rate=odpr)))
f_target = lambda x: proj(encoder(x, drop_path_rate=tdpr))

for v1, v2 in loader: # load two views
    q1, q2 = f_online(v1), f_online(v2) # queries: NxC
    k1, k2 = f_target(v1), f_target(v2) # keys: NxC

    # symmetric loss
    loss = contrastive_loss(q1, k2, queue2) + contrastive_loss(q2, k1, queue1)
    loss.backward()

    update(f_online) # optimizer update: f_online
    f_target = m * f_target + (1. - m) * f_online # momentum update: f_target
    update(m) # update momentum coefficient

def contrastive_loss(q, k, queue):
    # positive logits: Nx1
    l_pos = torch.einsum('nc,nc->n', [q, k.detach()]).unsqueeze(-1)
    # negative logits: NxK
    l_neg = torch.einsum('nc,ck->nk', [q, queue.clone().detach()])

    # logits: Nx(1+K)
    logits = torch.cat([l_pos, l_neg], dim=1)

    # labels: positive key indicators
    labels = torch.zeros(N)
    loss = F.cross_entropy(logits / t, labels)
    
    # update queue
    enqueue(queue, k)
    dequeue(queue)
    
    return loss
\end{lstlisting}
\end{algorithm}
\end{figure}

\paragraph{Swin Transformer as the backbone}

Swin Transformer is a general-purpose backbone for computer vision and achieved state-of-the-art performance on various vision tasks such as COCO object detection (58.7 box AP and 51.1 mask AP on test-dev set) and ADE20K semantic segmentation (53.5 mIoU on validation set). It is basically a hierarchical Transformer whose representation is computed with shifted windows. The shifted windowing scheme brings greater efficiency by limiting self-attention computation to non-overlapping local windows while also allowing for cross-window connection.

In this work, we adopt the tiny version of Swin Transformer (Swin-T) as our default backbone, such that the transferring performance on downstream tasks of object detection and semantic segmentation can be also evaluated. The Swin-T has similar complexity with ResNet-50 and DeiT-S. The details of specific architecture design and hyper-parameters can be found in~\citep{swin}.

\section{Experiments}

\subsection{Linear Evaluation on ImageNet-1K}
Linear evaluation on ImageNet-1K dataset is a common evaluation protocol to assess the quality of learnt representations~\citep{moco}. In this protocol, a linear classifier is applied on the backbone, with the backbone weights frozen and only the linear classifier trained. After training this linear classifier, the top-1 accuracy using center crop is reported on the validation set.

During training, we follow~\citep{moco} to use random resize cropping with scale from $[0.08, 1]$ and horizontal flipping as the data augmentation. 100-epoch training with a 5-epoch linear warm-up stage is conducted. The weight decay is set as 0. The learning rate is set as the optimal one of $\{0.5, 0.75, 1.0, 1.25\}$ through grid search for each pre-trained model.

Table~\ref{tab-linear} listed the major results of pre-trained models using different self-supervised learning methods and backbone architectures.

\paragraph{Comparison with other SSL methods using Transformer architectures} Regarding previous methods such as MoCo v3~\citep{mocov3} and DINO~\citep{dino} adopt ViT/DeiT as their backbone architecture, we first report results of MoBY using DeiT-S~\citep{deit} for fair comparison with them. Under 300-epoch training, MoBY achieves 72.8\% top-1 accuracy, which is slightly better than MoCo v3 and DINO (without the multi-crop trick), as shown in Table~\ref{tab-linear}. 

We note that MoCo v3 and DINO adopt heavy tricks to achieve the same accuracy as ours:
\begin{itemize}
    \item \emph{Tricks in MoCo v3~\citep{mocov3}}. MoCo v3 adopts a fixed patch embedding, batch normalization layers to replace the layer normalization ones before the MLP blocks, and a 3-layer MLP head. It also uses large batch size (i.e. 4096) which is unaffordable for many research labs. 
    \item \emph{Tricks in DINO~\citep{dino}}. DINO adopts asymmetric temperatures between student and teacher, a linearly warmed-up teacher temperature, varying weight decay during pre-training, the last layer fixed at the first epoch, tuning whether to put weight normalization in the head, a concatenation of the last few blocks or CLS tokens as the input to the linear classifier, and etc.
\end{itemize}

In contrast, we mainly adopt standard settings from MoCo v2~\citep{mocov2} and BYOL~\citep{byol}, and use a small batch size of 512 such that the experimental settings will be affordable for most labs. We have also started to try applying some tricks of MoCo v3~\citep{mocov3}/DINO~\citep{dino} to MoBY, though they are not included in the standard settings. Our initial exploration reveals that the fixed patch embedding has no use to MoBY, and replacing the layer normalization layers before the MLP blocks by batch normalization can bring +1.1\% top-1 accuracy using 100-epoch training, as shown in Table~\ref{tab-ablation-moco-v3}. This indicates that some of these tricks may be useful for the MoBY approach, and the MoBY approach has potential to achieve much higher accuracy on ImageNet-1K linear evaluation. This will be left as our future study.

\paragraph{Swin-T \textit{v.s.} DeiT-S} We also compare the use of different Transformer architectures in self-supervised learning. As shown in Table~\ref{tab-linear}, Swin-T achieves 75.0\% top-1 accuracy, surpassing DeiT-S by +2.2\%. Also note the performance gap is larger than that of using supervised learning (+1.5\%).

\begin{table}[h]
  \centering
  \begin{tabular}{ccccccc}
    \toprule
    Method & Arch. & Epochs & Params (M) & FLOPs (G) & img/s 
    & Top-1 acc (\%) \\
    \midrule
    Sup. & DeiT-S & 300 & 22 & 4.6 & 940.4 & 79.8 \\
    Sup. & Swin-T & 300 & 29 & 4.5 & 755.2 & 81.3 \\
    \midrule
    MoCo v3 & DeiT-S & 300 & 22 & 4.6 & 940.4 & 72.5 \\
    DINO & DeiT-S & 300 & 22 & 4.6 & 940.4 & 72.5 \\ 
    DINO$^{\dag}$ & DeiT-S & 300 & 22 & 4.6 & 940.4 & 75.9 \\
    \midrule
    MoBY & DeiT-S & 300 & 22 & 4.6 & 940.4 & 72.8 \\
    MoBY & Swin-T & 100 & 29 & 4.5 & 755.2 & 70.9 \\
    MoBY & Swin-T & 300 & 29 & 4.5 & 755.2 & \textbf{75.0} \\
    \bottomrule
  \end{tabular}
  \caption{Comparison of different SSL methods and different Transformer architectures on ImageNet-1K linear evaluation. $^{\dag}$ denotes DINO with a multi-crop scheme in training.}
  \label{tab-linear}
\end{table}

\begin{table}[h]
  \centering
  \begin{tabular}{ccc}
    \toprule
    Fixed Patch Embedding & Replace LN before MLP with BN & Top-1 acc (\%) \\
    \midrule
     &  & 70.9 \\
     \checkmark &  & 70.8 \\
     & \checkmark & 72.0 \\
    \bottomrule
  \end{tabular}
  \caption{Initial study of applying tricks in MoCo v3 to the MoBY approach using 100-epoch training and Swin-T backbone architecture. Note although replacing the layer norm layer before each MLP block with a batch norm layer performs better (72.0 vs. 70.9), it changes the original Swin architecture and is currently not used as our standard settings in experiments. We leave more comprehensive study of Transformer architecture improvements in the context of SSL as our future work.}
  \label{tab-ablation-moco-v3}
\end{table}

\subsection{Transferring Performance on Downstream Tasks}
We evaluate the transferring performance of the learnt representation on downstream tasks of COCO object detection/instance segmentation and ADE20K semantic segmentation.

\paragraph{COCO object detection and instance segmentation}
Two detectors are adopted in the evaluation: Mask R-CNN~\citep{maskrcnn} and Cascade Mask R-CNN~\citep{cascade}, following the implementation of~\citep{swin}\footnote{\url{https://github.com/SwinTransformer/Swin-Transformer-Object-Detection}}. Table~\ref{tab-coco} shows the comparison of the learnt representation by MoBY and the pretrained supervised method in~\citep{swin}, in both \texttt{1x} and \texttt{3x} settings. For each experiment, we follow all the settings used for supervised pre-trained models~\citep{swin}, except that we tune the drop path rate in $\{0, 0.1, 0.2\}$ and report the best results (for also supervised models).

It can be seen that the representations learnt by the self-supervised method (MoBY) and the supervised method are similarly well on transferring performance. While we note that previous SSL works using ResNet as the backbone architecture usually report stronger performance over the supervised methods~\citep{moco,pixpro,henaff2021efficient}, no gains over supervised methods are observed using Transformer architectures. We hypothesis it is partly because the supervised pre-training on Transformers has involved strong data augmentations~\citep{deit,swin}, while supervised training of ResNet usually employs much weaker data augmentation. These results also imply space to improve for self-supervised learning using Transformer architectures.

\begin{table}[h]
  \centering
  \begin{tabular}{ccccccccc}
    \toprule
    \multirowcell{2}{Method} & \multirowcell{2}{Model} & \multirowcell{2}{Schd.} 
    & \multicolumn{3}{c}{box AP} & \multicolumn{3}{c}{mask AP} \\
    \cmidrule(lr){4-6} \cmidrule(lr){7-9}
    & &
    & mAP$^{\text{bbox}}$ & AP$^{\text{bbox}}_\text{50}$ & AP$^{\text{bbox}}_\text{75}$
    & mAP$^{\text{mask}}$ & AP$^{\text{mask}}_\text{50}$ & AP$^{\text{mask}}_\text{75}$ \\
    \midrule
    \multirowcell{4}{Swin-T\\(mask R-CNN)} 
    & Sup. & 1x & 43.7 & 66.6 & 47.7 & 39.8 & 63.3 & 42.7 \\
    & MoBY & 1x & 43.6 & 66.2 & 47.7 & 39.6 & 62.9 & 42.2 \\
    \cmidrule{2-9}
    & Sup. & 3x & 46.0 & 68.1 & 50.3 & 41.6 & 65.1 & 44.9 \\
    & MoBY & 3x & 46.0 & 67.8 & 50.6 & 41.7 & 65.0 & 44.7 \\
    \midrule
    \multirowcell{4}{Swin-T\\(Cascade\\mask R-CNN)} 
    & Sup. & 1x & 48.1 & 67.1 & 52.2 & 41.7 & 64.4 & 45.0 \\
    & MoBY & 1x & 48.1 & 67.1 & 52.1 & 41.5 & 64.0 & 44.7 \\
    \cmidrule{2-9}
    & Sup. & 3x & 50.4 & 69.2 & 54.7 & 43.7 & 66.6 & 47.3 \\
    & MoBY & 3x & 50.2 & 68.8 & 54.7 & 43.5 & 66.1 & 46.9 \\
    \bottomrule
  \end{tabular}
  \caption{Comparison of the supervised method by ImageNet-1K classification and the self-supervised MoBY approach on transferring performance to COCO object detection and instance segmentation.}
  \label{tab-coco}
\end{table}

\paragraph{ADE20K Semantic Segmentation} The UPerNet approach~\citep{upernet} and the ADE20K dataset are adopted in the evaluation, following~\citep{swin}~\footnote{\url{https://github.com/SwinTransformer/Swin-Transformer-Semantic-Segmentation}}. The fine-tuning and testing settings also follow~\citep{swin} except that the learning rate of each experiment is tuned using $\{3\times 10^{-5}, 6\times 10^{-5}, 1\times 10^{-4}\}$. Table~\ref{tab-ade20k} shows the comparisons of supervised and self-supervised pre-trained models on this evaluation. It indicates that MoBY performs slightly worse than the supervised method, implying a space to improve for self-supervised learning using Transformer architectures.

\begin{table}[h]
  \centering
  \begin{tabular}{cccc}
    \toprule
    Method & Model & Schd.
    & mIoU \\
    \midrule
    \multirowcell{4}{Swin-T\\(UPerNet)}
    & Sup. & 160K & 44.51 \\
    & MoBY & 160K & 44.06 \\
    \cmidrule{2-4}
    & Sup.$^{\dag}$ & 160K & 45.81 \\
    & MoBY$^{\dag}$ & 160K & 45.58 \\
    \bottomrule
  \end{tabular}
  \caption{Comparison of the supervised method by ImageNet-1K classification and the self-supervised MoBY approach on transferring performance to ADE20K semantic segmentation. $^{\dag}$ denotes the results with multi-scale testing techniques.}
  \label{tab-ade20k}
\end{table}

\subsection{Ablation Study}

We perform ablation study using the ImageNet-1K linear evaluation protocol. Swin-T is used as the backbone architecture. In each ablation, we vary one hyper-parameter and other hyper-parameters are set as the default ones.

\begin{table}[h]
  \centering
  \begin{tabular}{cccc}
    \toprule
    Epochs & \emph{Online} dpr & \emph{Target} dpr & Top-1 acc (\%) \\
    \midrule
    100 & 0.05 & 0.0 & 70.9 \\
    100 & 0.1  & 0.0 & 70.9 \\
    100 & 0.2  & 0.0 & 70.9 \\
    100 & 0.1  & 0.1 & 69.0 \\
    \midrule
    300 & 0.05 & 0.0 & 74.2 \\
    300 & 0.1  & 0.0 & 75.0 \\
    300 & 0.2  & 0.0 & 75.0 \\
    \bottomrule
  \end{tabular}
  \caption{Ablation study on the drop path rates of \emph{online} and \emph{target} encoders.}
  \label{tab-ablation-droppath}
\end{table}

\paragraph{Asymmetric drop path rates is beneficial} Drop path has proved a useful regularization for supervised representation learning using the image classification task and Transformer architectures~\citep{deit,swin}. We also ablate the effect of this regularization in Table~\ref{tab-ablation-droppath}. Increasing the drop path regularization from 0.05 to 0.1 to the \emph{online} encoder is beneficial for representation learning, especially in longer training, probably due to the relief of over-fitting. Additionally adding drop path regularization to the \emph{target} encoder results in 1.9\% top-1 accuracy drop (70.9\% to 69.0\%), indicating a harm. We thus adopt an asymmetric drop path rates in pre-training.

\paragraph{Other hyper-parameters} Table~\ref{tab-ablation-queue} ablates the effect of \emph{key} queue size $K$ from 1024 to 16384. The approach stably performs across various $K$ (from 1024 to 16384), and we adopt 4096 as default.
Table~\ref{tab-ablation-temp} ablates the effect of temperature $\tau$ and 0.2 performs best which is set as the default value.
Table~\ref{tab-ablation-momentum} ablates the effect of the starting momentum value of the \emph{target} encoder. 0.99 performs best and is set as the default value.

\begin{table}[h]
  \begin{subtable}[h]{0.3\textwidth}
  \centering
  \begin{tabular}{cc}
    \toprule
    $K$ &  Top-1 (\%) \\
    \midrule
     1024 & 71.0 \\
     2048 & 70.8 \\
     4096$^*$ & 70.9 \\
     8192 & 71.0 \\
     16384 & 70.8 \\
    \bottomrule
  \end{tabular}
  \caption{Queue Size $K$}
  \label{tab-ablation-queue}
  \end{subtable}
  \hfill
  \begin{subtable}[h]{0.3\textwidth}
  \centering
  \begin{tabular}{cc}
    \toprule
    $\tau$ &  Top-1 (\%). \\
    \midrule
     0.07 & 62.7 \\
     0.1 & 67.7 \\
     0.2$^*$ & 70.9 \\
     0.3 & 70.8 \\
    \bottomrule
  \end{tabular}
  \caption{Temperature $\tau$}
  \label{tab-ablation-temp}
  \end{subtable}
  \hfill
  \begin{subtable}[h]{0.3\textwidth}
  \centering
  \begin{tabular}{cc}
    \toprule
    Start value &  Top-1 (\%) \\
    \midrule
     0.99$^*$ & 70.9 \\
     0.993 & 70.7 \\
     0.996 & 70.5 \\
     0.999 & 67.6 \\
    \bottomrule
  \end{tabular}
  \caption{Momentum of \emph{target} encoder}
  \label{tab-ablation-momentum}
  \end{subtable}
  \caption{Ablation study on other hyper-parameters using 100-epoch training. $^*$ denotes the default values.}
  \label{tab-ablation-hyper-params}
\end{table}

\section{Conclusion}

In this paper, we present a self-supervised learning approach called \textcolor{Maroon}{\texttt{MoBY}}, with Vision Transformers as its backbone architecture. With a proper training recipe and much lighter tricks than MoCo v3/DINO, MoBY can achieve reasonably high performance on ImageNet-1K linear evaluation: 72.8\% and 75.0\% top-1 accuracy using DeiT-S and Swin-T, respectively, by 300-epoch training. More importantly, in contrast to ViT/DeiT, the general-purpose Swin Transformer backbone enables us to also evaluate the learnt representations on downstream tasks such as object detection and semantic segmentation. MoBY can perform comparably or slightly worse than the supervised methods, indicating a space to improve for self-supervised learning with Transformer architectures. We hope our results can facilitate more comprehensive evaluation of self-supervised learning methods designed for Transformer architectures. Our code and models are available and will be continually enriched at \url{https://github.com/SwinTransformer/Transformer-SSL}.

\newpage

\bibliographystyle{apalike}
\bibliography{references}

\end{document}